\documentclass[conference,9pt]{IEEEtran}
\IEEEoverridecommandlockouts

\usepackage{cite}
\usepackage{amsmath,amssymb,amsfonts}
\usepackage{graphicx}
\usepackage{textcomp}
\usepackage{xcolor}
\usepackage{subcaption}
\usepackage{mathtools}
\usepackage{tipa}
\usepackage{censor}
\usepackage[utf8]{inputenc}
\usepackage{algorithm}
\usepackage{float}
\usepackage{algpseudocode}
\usepackage{url} 

\usepackage{siunitx}
\usepackage{xspace}
\def\BibTeX{{\rm B\kern-.05em{\sc i\kern-.025em b}\kern-.08em
    T\kern-.1667em\lower.7ex\hbox{E}\kern-.125emX}}

\def\F0{$F_0$\xspace}

\def\latents{\textsc{Latents}\xspace}
\def\latent{\textsc{Latent}\xspace}
\def\centroids{\textsc{Averaged Latents}\xspace}
\def\centroid{\textsc{Average Latent}\xspace}
\def\discretesymbols{\textsc{Discrete Symbols}\xspace}
\def\discretesymbol{\textsc{Discrete Symbol}\xspace}

\begin{document}

\title{Do Discrete Self-Supervised Representations of Speech Capture Tone Distinctions?\\

\thanks{This work was supported in part by the UKRI Centre for Doctoral Training in Natural Language Processing, funded by the UKRI (grant EP/S022481/1) and the University of Edinburgh, School of Informatics and School of Philosophy, Psychology \& Language Sciences.}
}


\author{
\IEEEauthorblockN{Opeyemi Osakuade and Simon King}
\IEEEauthorblockA{\textit{The Centre for Speech Technology Research}, 
\textit{University of Edinburgh}, UK \\
O.M.Osakuade@sms.ed.ac.uk, Simon.King@ed.ac.uk}
}

\maketitle

\begin{abstract}
Discrete representations of speech, obtained from Self-Supervised Learning (SSL) foundation models, are widely used, especially where there are limited data for the downstream task, such as for a low-resource language. Typically, discretization of speech into a sequence of symbols is achieved by unsupervised clustering of the latents from an SSL model. Our study evaluates whether discrete symbols - found using k-means - adequately capture tone in two example languages, Mandarin and Yoruba. We compare latent vectors with discrete symbols, obtained from HuBERT base, MandarinHuBERT, or XLS-R, for vowel and tone classification. We find that using discrete symbols leads to a substantial loss of tone information, even for language-specialised SSL models. We suggest that discretization needs to be task-aware, particularly for tone-dependent downstream tasks.


\end{abstract}

\section{Introduction}


Foundation models trained using Self-Supervised Learning (SSL) have become an important resource in spoken language modelling, and are particularly useful when dealing with challenging situations such as insufficient annotated data in the target language or domain. 

\subsection{Speech representations from Self-Supervised Learning}


The representations of speech learned by such models (henceforth ``SSL speech representations'') have led to improvements in many tasks \cite{yang21c_interspeech} including Automatic Speech Recognition \cite{irvin2023self}, language identification \cite{fan21_interspeech}, and speech-to-speech translation \cite{lee2022direct,zhang2021uwspeech,chen2023speech}. SSL speech representations excel at distinguishing between different phonetic classes \cite{wells2022phonetic,van2023rhythm}. For instance, wav2vec~2.0, \cite{baevski2020wav2vec} captures phonetic detail during pre-training on unlabelled audio data and its SSL speech representations achieve state-of-the-art performance on ASR tasks with only minimal
labelled data. Similarly, clustering the SSL speech representations from HuBERT \cite{hsu2021hubert} leads to the discovery of phone-like classes,
without any explicit annotation. SSL speech representations have also been shown to encode both speaker and phonetic information. Recent probing experiments have demonstrated the effectiveness of new methods in eliminating speaker information while simultaneously outperforming previous baselines in phone discrimination tasks \cite{liu2023self}. 

\subsection{The benefits of using discrete representations}


It is increasingly common to discretize SSL speech representations. For the remainder of this paper, we will use the following terms. The underlying SSL model generally provides continuously-valued vectors (e.g., the activations from layer 9 of a HuBERT model) called \latents. These can be discretized, typically by clustering, with each cluster  represented as a \discretesymbol from a closed vocabulary. The \discretesymbols are sometimes referred to  \textit{tokens}, by analogy with text tokens. It is also possible to average the sequence of \latents for each speech unit of interest (a phone, in our work), resulting in a single \centroid of the same dimensionality as a \latent.

The authors of \cite{van2017neural} argue that requiring their VQ-VAE model to use a quantized  representation forces it to capture more robust and meaningful information, reducing unnecessary variability. Sometimes discretization occurs during model training, but it is also common to quantize \latents from an \textit{already-trained} model using a simple task-agnostic clustering technique such as k-means.

In addition to the benefits of compression (e.g., for storage or transmission), discretizing continuous vectors makes them behave like text tokens, opening up the direct application of natural language processing (NLP) techniques -- notably language models -- to speech tasks. 

Once speech has been transformed into a sequence of \discretesymbols, they can be readily mixed (e.g., by taking the union of the vocabularies) with symbols from other modalities to perform multimodal tasks across image, audio, video, and text\cite{anil2023gemini}. \cite{baevski2019effectiveness} found \discretesymbols from the vq-wav2vec model \cite{Baevski2020vq-wav2vec} to be accurate for NLP tasks. \cite{lee2022direct} showed the possibilities for direct speech-to-speech translation (S2ST) using \discretesymbols for source and target speech without the use of text.
\discretesymbols of the target speech have been used in S2ST for unwritten languages such as Hokkien \cite{chen2023speech}.
Representing speech as \discretesymbols also simplifies audio generation tasks, such as speech enhancement or synthesis, because this converts the task into classification, rather than complex, high-dimensional regression \cite{mousavi2024should}.

In summary, there are well-motivated reasons to represent speech as a sequence of \discretesymbols. There is a strong correlation between \discretesymbols and phonetic class, and  broad phonetic classes, but a weaker correlation with speaker characteristics such as gender \cite{sicherman2023analysing}. \discretesymbols also capture sub-phonetic dynamics such as the distinct closure and release phases of plosive consonants \cite{wells2022phonetic}, suggesting that sequences of \discretesymbols are capable of capturing the fine-grained details of speech production, offering a rich representation of speech suitable for myriad downstream tasks.

\subsection{Potential downsides of discrete representations}




While discretization can effectively filter out non-linguistic features like background noise or speaker identity, there are trade-offs, most obviously in  selecting the optimal codebook (i.e., vocabulary) size:  large enough to capture the required speech detail but otherwise as small as possible to facilitate subsequent (language) modelling. Our concern in the current work is the potential loss of \F0 (pitch) speech characteristics, which are essential for downstream tasks involving intonation or tone.

In the literature, discretization appears to be most commonly task-agnostic, such as k-means. As we will see later, this might be particularly prone to the loss of task-specific information. 



\subsection{Contributions of our work}

We investigate whether \discretesymbols -- derived using k-means from popular SSL models -- capture \textbf{tone}. We use two different approaches to probe for tone, in two example languages, Mandarin and Yoruba.







\section{Tone languages}
Tones are a fundamental aspect of many languages, with an estimated 60\% - 70\% of the world's languages  presumed to use tone, \cite{zhiming2003moira} particularly those within the linguistic families of East Asia, Africa, and South America. Tone is the use of pitch  (\F0) variations in addition to vowels and constants to distinguish words with different meanings that would otherwise be homophones \cite{best2019diversity}; this is also called \textit{lexical tone} \cite{singh2016new}. While many (non-tone) languages may use pitch to convey meaning through intonation, tone languages use pitch variations to distinguish between lexical words \cite{li2021human}.

Tones are patterns of pitch variation \cite[Chapter 29]{crystal2010}. The number and shape of tones differs from one language to another; for instance, Mandarin Chinese has 4 main tones (high, rising, falling-rising, falling), Yoruba has 3 (rising, falling, neutral), whereas some African languages such as Zulu possess more complex tone systems. In Mandarin Chinese, the most widely-spoken tone language, the word "ma" can mean mother", "hemp", "horse", or "scold" depending on which of the four tones is used \cite{wang2013perception}.
In Yoruba, spoken in Nigeria and neighbouring countries, the three tones not only distinguish individual words but also convey grammatical structure. Tones are marked in the standard orthography using diacritics on vowels and nasals \cite{oyelaran1971yoruba}. For example, the word ``ogun'' can mean `war', `inheritance', `twenty', or `medicine' depending on the tone used. Tables \ref{YorubA} and \ref{Mandarin} show the  vowels and tones for each language that will be analysed in our experiments.
Yoruba has nasal vowels, which we do not include in this analysis.
\begin{table}[th]
\centering
\caption{Yoruba vowel and tone inventory (Epitrans notation)}
\begin{tabular}{c|c}
\hline
\textbf{Category} & \textbf{Classes} \\ \hline
vowel-with-tone & \begin{tabular}[c]{@{}c@{}}a, aH, aL, e, eH, eL, \textepsilon, {\textepsilon}H
,\\ {\textepsilon}L, i, iH, iL, o, oH, oL, \\ \textopeno, {\textopeno}H, {\textopeno}L, u, uH, uL \end{tabular} \\ \hline
vowel-without-tone & a, e, \textepsilon, i, o, \textopeno, u \\ \hline
tone-only & \begin{tabular}[c]{@{}c@{}}High (H), Low (L), Neutral ( )\end{tabular} \\ \hline
\end{tabular}
\label{YorubA}
\end{table}

Mandarin has other vowels, including diphthongs, which for simplicity we do not include in the current analysis.
\begin{table}[th]
\centering
\caption{Mandarin monophthong vowel and tone inventory (AISHELL notation)}
\begin{tabular}{c|c}
\hline
\textbf{Category} & \textbf{Classes} \\ \hline
vowel-with-tone & \begin{tabular}[c]{@{}c@{}}a1, a2, a3, a4, a5, e1, e2, e3, e4,\\ e5, i1, i2, i3, i4, i5, o1, o2, o3, \\ o4, o5, u1, u2, u3, u4, u5\end{tabular} \\ \hline
vowel-without-tone & a, e, i, o, u \\ \hline
tone-only & \begin{tabular}[c]{@{}c@{}}flat (1), rising (2), dip (3), \\ falling (4), neutral (5) \end{tabular} \\ \hline
\end{tabular}
\label{Mandarin}
\end{table}

\section{Related work in SSL speech representations for tone languages}
Tones are suprasegmental features that spread across multiple phones and are subject to coarticulatory effects \cite{chen2022computational}. Recent research \cite{shen2024encoding} has shown that speech language models trained on wav2vec 2.0 \latents  encode lexical tone information for Mandarin and Vietnamese to a significant degree, regardless of whether they are trained on tone or non-tone languages. However, Chinese speech synthesis based on \discretesymbols \cite{tao2024toneunit} exhibited ``tone shift'': synthesized speech contained the correct base syllables but incorrect tones. To address this, the authors introduced a model-specific speech discretization framework to generate tone-aware speech units for speech synthesis. While this approach showed some improvement, it is bespoke to one model and relies on additional supervision from tone-labeled text. In other work, \discretesymbols have been used in speech-to-speech translation for the unwritten language Hokkien, but that system required additional supervision from Mandarin text during training to provide more information about tone \cite{chen2023speech}.

Our novel contribution in the remainder of this paper will be to demonstrate that the tone problems encountered by the above work are caused \textit{solely} by quantization, rather than elsewhere in the system.

SSL speech representations can be \textbf{probed} to determine whether important linguistic information, such as phonetic or tone distinctions, is preserved in latents, centroids, or tokens. There are numerous probing techniques in the literature, one well-established method involves classification tasks\cite{choi2024self,ma2021probing}. Another is the ABX task, often used to assess the discriminability of phonetic distinctions in SSL representations by comparing pairs of audio segments based on specific phonetic features \cite{martin23_interspeech}. We use two techniques in the current work. The first is a tried-and-tested classification task. The second approach is novel and involves measuring the average distance between all pairs of tone-carrying vowel phones in a corpus, for each language separately. Note that we use the term phone consistently to refer to an individual spoken realisation.






\section{Methodology}

We utilized Mandarin Chinese data from AISHELL-1 \cite{bu2017aishell} and Yoruba data from BibleTTS \cite{meyer2022bibletts}. AISHELL-1 consists of over \qty{170}{\hour} hours of \qty{16}{\kHz} recordings from 400 speakers, while the Yoruba corpus contains 93 hours of studio-quality, \qty{48}{\kHz} recordings from a single speaker. All audio was downsampled to \qty{16}{\kHz} / \qty{16}{\bit}, as required by these models. We extracted \latents from the 9th layer of: HuBERT base model trained on English data from scratch and XLS-R (a wav2vec~2.0 model fine-tuned on 128 languages including $\sim$22 tone languages including Yoruba and Mandarin) from fairseq\footnote{\url{https://github.com/facebookresearch/fairseq/tree/main/examples/textless_nlp/gslm/speech2unit}}, and MandarinHuBERT from Hugging Face\footnote{\url{https://huggingface.co/TencentGameMate/chinese-hubert-base}}. The 9th layer is known to capture linguistic information \cite{pasad2021layer}. The \latent dimension is 768 or 1024, for HuBERT or XLS-R respectively.

Phonetic forced alignments were obtained using the Montreal Forced Aligner \cite{mcauliffe2017montreal}, using pronunciations provided by Epitran grapheme-to-phone\cite{mortensen2018epitran}. Given these phone alignments, \latents were averaged within each phone in the corpus to obtain the \centroids.

The \latents were k-means clustered into $k$ clusters with $K = 50, 100, 200$, separately for each language, following the approach outlined in \cite{lee2022direct}, increasing $K$ to $1000$ does not make much difference. This clustering is therefore corpus-specific but task-agnostic. We chose a few values for k that are in the  range found in the literature but only present results for $K = 200$ .

The resulting  \latents, \centroids, and \discretesymbols are passed to the two probing approaches explained in the following Sections.

\subsection{Probing using a classification task}

The most obvious way to probe for the presence of tone is to try to classify it. \latents and \discretesymbols are both sequences, for which we trained Long Short-Term Memory (LSTM) classifiers. \centroid is a single vector per phone, for which we used Logistic Regression (LR). The data were divided into an 80:20 train-test split and we trained a total of 9 classifiers per language: 3 representations (\latents, \centroids, \discretesymbols) $\times$ 3 classification tasks:
\begin{itemize}
\item \textbf{vowel-with-tone}: vowel phones retaining their tone label
\item \textbf{vowel-without-tone}: vowel phones only, ignoring tone 
\item \textbf{tone-only}: vowel phones, ignoring phone class
\end{itemize}

\subsection{Probing using pairwise edit distance}\label{edit_distance}
Our second probing technique is novel, and only applies to \discretesymbols. Each phone is represented as a sequence of \discretesymbols, varying in length according to the duration in frames of that phone (on average, around 5-10 \discretesymbols per phone). We use edit (Levenshtein) distance to measure the distance between every possible pair of vowel phones in the corpus. Edit distance is the smallest number of insertions, deletions, or substitutions required to transform one sequence into the other, and is found efficiently using Dynamic Programming \cite{wagner1974string}.
Edit distance has proved effective in recent studies of phonetic variability, to analyze phonetic sequences in unsupervised or semi-supervised learning settings \cite{dunbar2020zero,baevski2020wav2vec}.
Our motivation for using it is that tone \textit{might} be encoded in patterns of symbol sequences, but not in individual symbols.

Averaging these pairwise distances across phones with particular properties enables us to evaluate how well the discrete acoustic units capture phonetic and tone distinctions. An example, for the vowel-without-tone condition, will best explain our method: forming all possible pairs of \textipa{[a]} phones and \textipa{[E]} phones in the corpus, measuring the pairwise edit distances, then taking the average, will tell us how well the \discretesymbol representation preserves that phonetic distinction. This distance can be visualised as one cell in a distance matrix along with all other pairings.

In the distance matrices presented in Figures \ref{fig:hu_man_vowels} and \ref{fig:hu_yor_vowels}, lighter shades indicate lower distance and darker shades represent higher distance. If an SSL model captures a distinction well, we should see lighter shades along the diagonal (self-similarity of phones from the same class) and darker shades elsewhere.

Given the results of the classification probe, our hypothesis is that tone will be less well captured than phonetic class, also aligning with recent findings in speech representation learning, where phonetic distinctions tend to be captured better by self-supervised models than prosodic or tone features, especially in tone languages \cite{lakhotia2021generative}. 

\section{Results and analysis}




\begin{table*}[htbp]
\caption{Mandarin: classification F1 scores}

\begin{center}
\begin{tabular}{|r|c|c|c|c|c|c|}
\hline
\multicolumn{1}{|r|}{Model$\rightarrow$} & \multicolumn{3}{c|}{\textbf{HuBERT base}} & \multicolumn{3}{c|}{\textbf{MandarinHuBERT}} \\ \hline
\multicolumn{1}{|r|}{Representation$\rightarrow$} & \latents & \centroids & \discretesymbols & \latents & \centroids & \discretesymbols \\ \hline
\textbf{\begin{tabular}[c]{@{}c@{}}vowel-without-tone\end{tabular}} & 0.97 & 0.94 & 0.79 & 0.99 & 0.98 & 0.86 \\ \hline
\textbf{\begin{tabular}[c]{@{}c@{}}vowel-with-tone\end{tabular}} & 0.70 & 0.62 & 0.38 & 0.79 & 0.74 & 0.46 \\ \hline
\textbf{tone} & 0.71 & 0.65 & 0.45 & 0.79 & 0.76 & 0.49 \\ \hline
\end{tabular}
\end{center}
\label{tab:combined_mandarin}
\end{table*}

\begin{table*}[htbp]
\caption{Yoruba: classification F1 scores}

\begin{center}
\begin{tabular}{|r|c|c|c|c|c|c|}
\hline
\multicolumn{1}{|r|}{Model$\rightarrow$} & \multicolumn{3}{c|}{\textbf{HuBERT base}} & \multicolumn{3}{c|}{\textbf{XLS-R}} \\ \hline
\multicolumn{1}{|r|}{Representation$\rightarrow$}  & \latents & \centroids & \discretesymbols & \latents & \centroids & \discretesymbols \\ \hline
\textbf{\begin{tabular}[c]{@{}c@{}}vowel-without-tone\end{tabular}} & 0.96 & 0.92 & 0.57 & 0.97 & 0.96 & 0.60 \\ \hline
\textbf{\begin{tabular}[c]{@{}c@{}}vowel-with-tone\end{tabular}} & 0.83 & 0.78 & 0.33 & 0.65 & 0.86 & 0.37 \\ \hline
\textbf{tone-only} & 0.86 & 0.74 & 0.49 & 0.89 & 0.82 & 0.52 \\ \hline
\end{tabular}
\end{center}
\label{tab:combined_yoruba}
\end{table*}


\subsection{Results for the classification probe}

Table \ref{tab:combined_mandarin} and \ref{tab:combined_yoruba} present the results of the classification probe, reported as F1 rather than accuracy, because vowel and tone distributions are highly uneven.

\subsubsection{Analysis by classification task}

For the hardest task (largest number of classes to distinguish) of vowel-with-tone, F1 scores are generally lowest, yet the F1 scores for tone-only task (with only 5 or 3 classes for Mandarin or Yoruba respectively) are little better. We can already conclude that, regardless of language, SSL model, or representation, tone classification is harder that phone classification.

\subsubsection{Analysis by representation}

For both languages and both SSL models, F1 for the classification tasks involving tone (the final 2 rows of the table) declines a little from \latents to \centroids, but then drops \textbf{very substantially} for \discretesymbols. We can conclude that the continuous representations yield better performance (even when averaged to one vector per phones), but that discretization removes a substantial amount of tone information. The vowel-without-tone task exhibits a less severe performance drop when moving from the continuous representations to the discrete one.

\begin{itemize}
\item \textbf{\latents}: 
Unsurprisingly, this consistently achieves the highest classification accuracy for all languages, models, and tasks. For the task of vowel-without-tone: in Mandarin, HuBERT base \latents achieve 0.97, while MandarinHuBERT improves this to 0.99. In Yoruba, HuBERT base \latents achieve 0.96, with XLS-R slightly higher at 0.97. 



\item \textbf{\centroids}: because \latents have a long sequence length (typically around 50 vectors per second), it is common in many downstream tasks to take an average over each linguistic unit of interest (here a phone, but could be a word, etc) to dramatically reduce the sequence length. As expected, this generally reduces F1 for our simple classification probe.\footnote{There is one anomalous result in Table \ref{tab:combined_yoruba} of 0.65 for XLS-R \latents on the vowel-with-tone task, which we will investigate in future work.} For example: in Mandarin, vowel-with-tone accuracy drops from 0.70 to 0.62 (HuBERT base) and from 0.79 to 0.74 (MandarinHuBERT). Yoruba follows a similar pattern.

\item \textbf{\discretesymbols}:
This representation has generally lower performance across the board, but is especially poor for the two tasks requiring tone classification, regardless of language or SSL model. 


\end{itemize}

Overall, 

\subsubsection{Analysis by language and SSL model}
The patterns are similar for both languages and both SSL models.

The specialised MandarinHuBERT model performs slightly better than HuBERT base for Mandarin. For vowel-without-tone, \latents from MandarinHuBERT achieve a near-perfect 0.99. But even HuBERT base (trained only on English) provides 0.97. Likewise for Yoruba, the multilingual XLS-R model provides excellent vowel-without-tone classification of 0.97, with English-only HuBERT base nearly as good, at 0.96. We conclude that language-specific or multilingual models are not essential for vowel classification.

Where we might expect those SSL models to do better is on tone classification. However, inspection of the last two rows of Tables \ref{tab:combined_mandarin} and \ref{tab:combined_yoruba} reveals very limited improvements when replacing HuBERT base with a language-specialised model.




The overall conclusion from the classification probe is clear: all models perform vowel classification very well, do less well with tone, and suffer a dramatic reduction when moving from a continuous representation to a discrete one.

\subsection{Results for the pairwise distance probe}

\begin{figure*}[h!]
  \centering
  \hspace{0.01cm} 
  \begin{subfigure}{9cm}
    \includegraphics[width=\linewidth]
    {corr_plot/man_plot.pdf}
    \caption{HuBERT base on Mandarin. (MandarinHuBERT plot not shown for reasons of space, but the pattern is similar.)}
    \label{fig:hu_man_vowels}
  \end{subfigure}
  \begin{subfigure}{8.5cm}
    \includegraphics[width=\linewidth]
    {corr_plot/yor_plot.pdf}
    \caption{HuBERT base on Yoruba. (XLS-R plot not shown for reasons of space, but the pattern is similar.)}
    \label{fig:hu_yor_vowels}
  \end{subfigure}
  \hspace{0.01cm} 
\end{figure*}




The pairwise edit distances for \discretesymbols described in Section \ref{edit_distance} are visualised in Figures \ref{fig:hu_man_vowels} and \ref{fig:hu_yor_vowels}. 
There is a clear global block pattern for Mandarin indicating that HuBERT base (and MandarinHuBERT, not plotted here) can discriminate between phonemes (classes of phones), consistent with the good classification results for the vowel-without-phone task above.


Figure \ref{fig:hu_yor_vowels} presents a slightly different pattern for HuBERT on Yoruba: limited success in distinguishing specific vowels but better on broad classes, such as high confusibility (low edit distances) within the \textipa{/e/}, \textipa{/E/}, \textipa{/i/} group, but good discrimination between this group and the other vowels. This is consistent with the very low classification performance for \discretesymbols in Table \ref{tab:combined_yoruba}.

Inspecting the sub-matrices for tone discrimination ($5 \times 5$ for Mandarin; $3 \times 3$ for Yoruba), we can see no obvious diagonal pattern. Again, this is consistent with the poor classification results for the two tasks involving tone, for both languages. We can conclude that there is no evidence of tone in patterns of symbol sequences. 

\section{Conclusion}



We have investigated whether representations of speech derived by Self-Supervised Learning (SSL) capture tone information, for two example tone languages: Mandarin and Yoruba.

Since the SSL model HuBERT is trained solely on English, we also included XLS-R (multilingual) and MandarinHuBERT (specialised to Mandarin). Both offered only modest gains over HuBERT at distinguishing tone.

To recap, the primary motivation for discretization is that language modelling techniques can be applied to speech tasks, with a secondary motivation that \discretesymbols  (more by luck than design) filter out unwanted non-linguistic features whilst retaining phonetic information. Unfortunately, it is now clear that important speech characteristics, notably tone, are also filtered out.


Of course, it would be possible to simply increase the number of clusters during k-means, leading to a larger vocabulary of \discretesymbols, which would -- by definition -- lose less information. But with increasing $K$ to $1000$ not making much difference, this is problematic for language modelling, where the smallest possible vocabulary is highly desirable.

Our main conclusion is that the solution to using \discretesymbols for tone languages lies \textit{not} in training (or finetuning) the SSL model with language-specific data, but rather in improving the discretization method. We employed what is perhaps the most common method: task-agnostic k-means. The obvious solution is some form of task-aware discretization that preserves the distinctions (e.g., tone) required by the downstream task. 
Future work could, for example, devise a tone-preserving discretization that would provide an elegant and general-purpose upstream solution to tone-related problems encountered in downstream tasks such as speech synthesis \cite{tao2024toneunit} or speech translation \cite{chen2023speech}.




\bibliographystyle{IEEEtran}
\bibliography{discrete_reps_SSL}

\end{document}